\documentclass[a4paper,12pt]{article}
\usepackage[utf8]{inputenc}
\usepackage{amsmath,amssymb,amsthm,graphicx,rotating,color,enumerate}

\newenvironment{dashlist}{
  \begin{enumerate}[--]
  \setlength{\itemsep}{0pt}
  \setlength{\parskip}{0pt}
  \setlength{\parsep}{0pt}
}{\end{enumerate}}

\DeclareMathSymbol{\leqslant}{\mathalpha}{AMSa}{"36}
\DeclareMathSymbol{\geqslant}{\mathalpha}{AMSa}{"3E}

\renewcommand{\ge}{\;\geqslant\;}
\def\RR{{\mathbb R}}

\def\qq{\qquad}
\def\q{\quad}

\title{Sparse arrays of signatures for online character recognition}
\author{
Benjamin Graham\\
{\small Dept of Statistics, University of Warwick, CV4 7AL, UK}\\
{\small \tt b.graham@warwick.ac.uk}\\
}

\begin{document}
\maketitle
\begin{abstract}
In mathematics the {\em signature} of a path is a collection of iterated integrals, commonly used for solving differential equations.
We show that the path signature, used as a set of features for consumption by a convolutional neural network (CNN), improves the accuracy of  {\em online} character recognition---that is the task of reading characters represented as a collection of paths.
Using  datasets of letters, numbers, Assamese and Chinese characters, we show that the first, second, and even the third iterated integrals contain useful information for consumption by a CNN.

On the CASIA-OLHWDB1.1 3755 Chinese character dataset, our approach gave a test error of 3.58\%, compared with 5.61\%\cite{multicolumndeep} for a traditional CNN. A CNN trained on the CASIA-OLHWDB1.0-1.2 datasets won the ICDAR2013 Online Isolated Chinese Character recognition competition.


Computationally, we have developed a sparse CNN implementation that make it practical to train CNNs with many layers of max-pooling. Extending the MNIST dataset by translations, our sparse CNN gets a test error of 0.31\%.

\noindent {\bf Keywords:} online character recognition, signature, iterated integrals, convolutional neural network\\
\noindent {\bf Condensed running title:} Sparse arrays of signatures
\end{abstract}

\section{Introduction}

Two rather different techniques work well for online Chinese character recognition.
One approach is to render the strokes into a $40\times40$ bitmap  embedded in a $48\times48$ grid, and then to use a deep convolutional neural network (CNN) as a classifier \cite{multicolumndeep}.
Another is to draw the character on an $8\times8$ grid, and then in each square calculate a {\em histogram} measuring the amount of movement in each of the 8 compass directions, producing a $512$-dimensional vector to classify \cite{bb104561}.

Intuitively, the first representation records more accurately {\em where} the pen went, while the second is better at recording the {\em direction} the pen was taking. We attempt to get the best of both worlds by producing an enhanced picture of the character using the path {\em iterated-integral signature}. This value-added picture of the character records the pen's location, direction and the forces that were acting on the pen as it moved.

CNNs start with an input layer of size $N\times N\times M$. The first two dimensions are spatial; the third dimension is simply a list of features available at each point; for example, $M=1$ for grayscale images and $M=3$ for color images. When calculating the path signature, we have a choice of how many iterated integrals to calculate. If we calculate the zeroth, first, second, \dots, up to the $m$-th iterated integrals, then the resulting input vectors are $M=1+2+2^2+\dots+2^m$ dimensional.

This representation is {\em sparse}. We only calculate path signatures where the pen actually went: for the majority of the $N\times N$ spatial locations, the $M$-dimensional input vector is simply taken as all-zeros. Taking advantage of this sparsity makes it practical to train much larger networks than would be practical with a traditional CNN implementation.

\section{Sparse CNNs: DeepCNet$(l,k)$}
Inspired by \cite{multicolumndeep}, we have considered a simple family of CNNs with alternating convolutional and max-pooling layers. Let DeepCNet$(l,k)$ denote the CNN with
\begin{dashlist}
\item an input layer of size $N\times N \times M$ where $N=3\cdot 2^l$,
\item $k$ convolutional filters of size $3\times 3$ in the first layer,
\item $nk$ convolutional filters of size $2\times 2 $ in layers $n=2,\dots,l$
\item a layer of $2\times2$ max-pooling after each convolution layer, and
\item a fully-connected final hidden layer of size $(l+1)k$.
\end{dashlist}
For example, DeepCNet$(4,100)$ is the architecture from \cite{multicolumndeep} with input layer size $N=48=3\cdot 2^4$ and four layers of max-pooling:
\[
\text{input-100C3-MP2-200C2-MP2-300C2-MP2-400C2-MP2-500N-output}
\]
For general input the cost of the forward operation, in particular calculating the first few hidden layers, is very high. For sparse input, the cost of calculating the lower hidden layers is much reduced, and evaluating the upper layers becomes the computational bottleneck.

When designing a CNN, it is important that the input field size $N$ is  strictly larger than the objects to be recognized; CNNs do a better job distinguishing features in the center of the input field. However, padding the input in this way is normally expensive. An interesting side effect of using sparsity is that the cost of padding the input disappears.

\subsection{Character scale $n\approx N/3$}
For character recognition, we choose a scale $n$ on which to draw the characters.
For the Latin alphabet and Arabic numerals, one might copy MNIST and take $n=20$. Chinese characters have a much higher level of detail: \cite{multicolumndeep} uses $n=40$, constrained by the computational complexity of evaluating dense CNNs.

\begin{figure}
\begin{center}
\includegraphics[width=6.5cm,trim=2.5cm 8cm 2.5cm 8cm,clip]{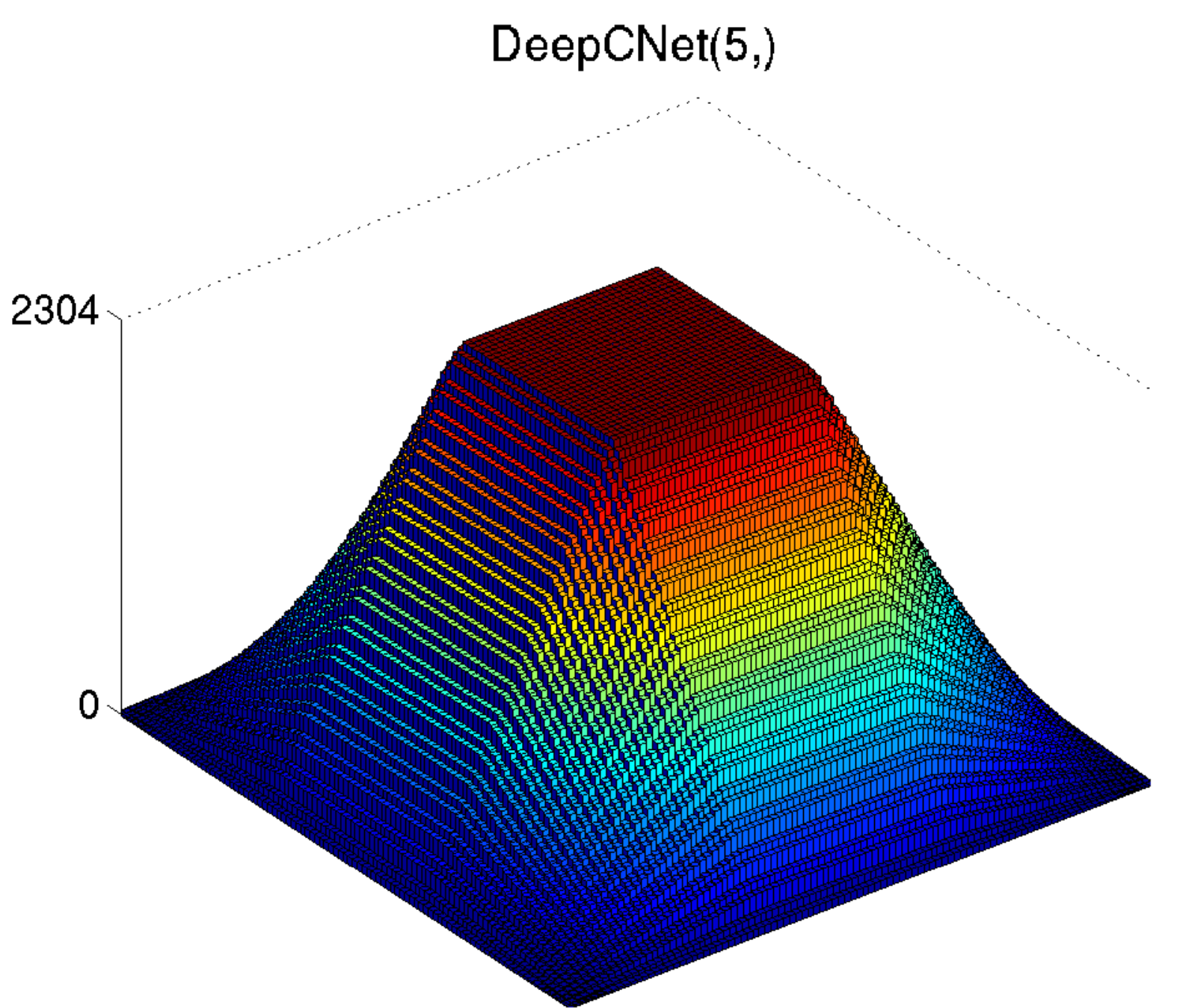}
\includegraphics[width=6.5cm,trim=2.5cm 8cm 2.5cm 8cm,clip]{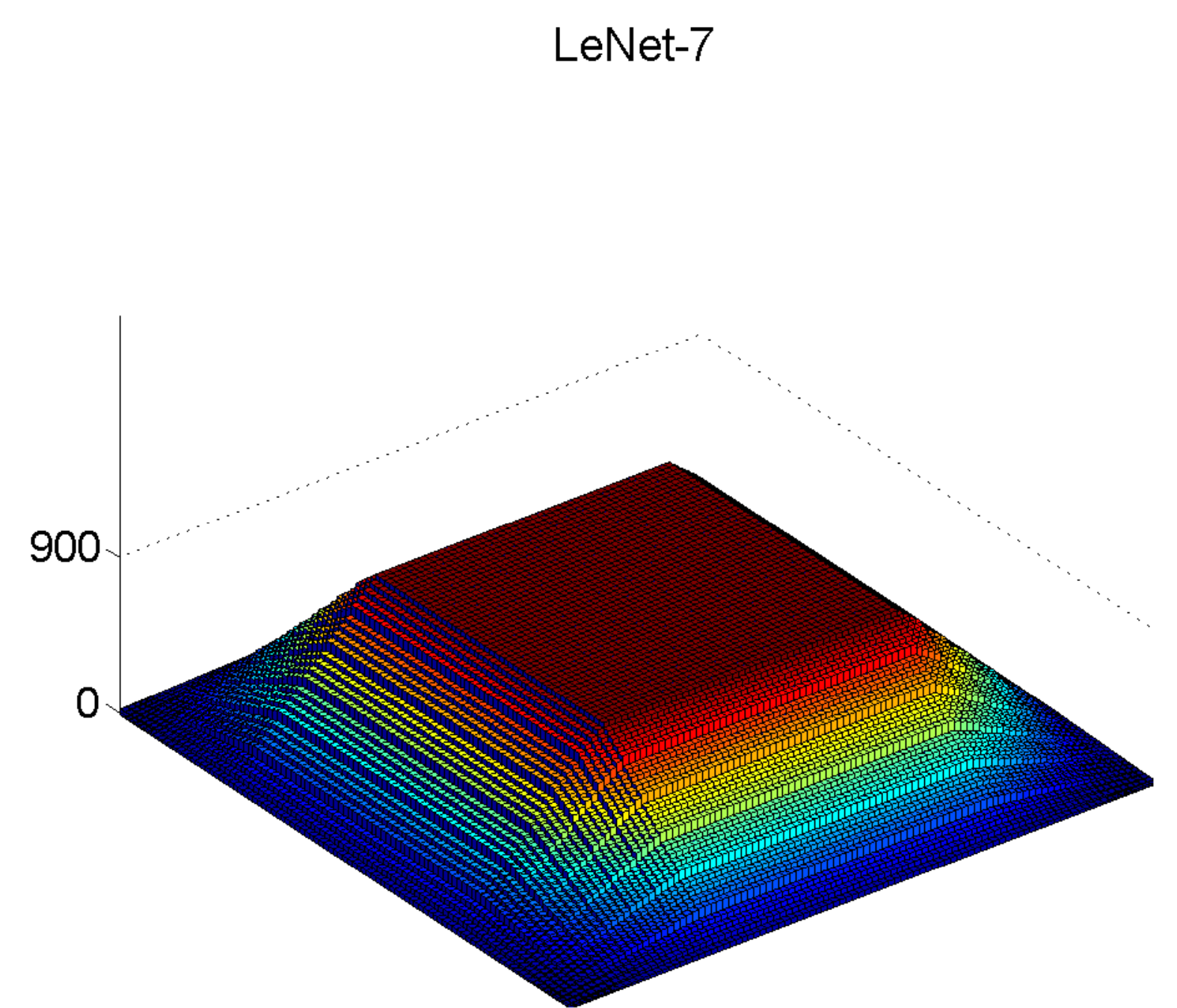}
\caption{A comparison of the number of possible paths from the input layer ($96\times 96$) to the fully connected layer for $l=5$ DeepCNets and LeNet-7 \cite{lecun-04}. The DeepCNet's larger number of hidden layers translates into a larger number of paths, but with a narrower plateau.\label{counts}}
\end{center}
\end{figure}

Given $n$, we must choose the $l$-parameter such that the characters fit comfortably into the input layer. DeepCNets seem to work best when $n$ is approximately $N/3$. There are a couple of ways of justifying the $n\approx N/3$ rule:
\begin{dashlist}
\item To process the $n\times n$ sized input down to a zero-dimensional quantity, the number of levels of $2\times2$ max-pooling $l=\log_2(N/3)$ should be approximately $\log_2 n$.
\item Counting the number of paths through the CNN from the input to output layers reveals a plateau; see Figure~\ref{counts}. Each corner of the input layer has only one route to the output layer; in contrast, the central $(N/3)\times(N/3)$ points in the input layer each have $9\cdot 4^{l-1}$ such paths.
\end{dashlist}

\subsection{Sparsity}
For DeepCNets with $l\ge 4$, training the network is in general hard work, even using a GPU.
However, for character recognition, we can speed things up by taking advantage of the sparse nature of the input, and the repetitive nature of CNN calculations. Essentially, we are memoizing the filtering and pooling operations.

First imagine putting an all-zero array into the input layer. As you evaluate the CNN in the forward direction, the translational invariance of the input is propagated to each of the hidden layers in turn. We can therefore think of each hidden variable as having a {\em ground state} corresponding to receiving no meaningful input; the ground state is generally non-zero because of bias terms. When the input array is sparse, one only has to calculate the values of the hidden variables where they differ from their ground state.

To forward propagate the network we calculate two types of list: lists of the non-ground-state vectors (which have size $M$, $k$, $2k$, \dots) and lists specifying where the vectors fit spatially in the CNN. This representation is very loosely biologically inspired. The human visual cortex separates into two streams of information: the dorsal ({\em where}) and ventral ({\em what}) streams.
\subsection{MNIST as a sparse dataset}
To test the sparse CNN implementation we used MNIST \cite{mnistlecun}. The $28\times28$ digits have on average 150 non-zero pixels.
Placing the digits in the middle of a $96\times 96$ grid produces a sparse dataset as $150$ is much smaller than $96^2$.

It is common to extend the MNIST training set by translations and/or elastic distortions. Here we only use translations of the training set, adding random shifts of up to $\pm2$ pixels in the $x$- and $y$-directions.  Training a very small-but-deep network, DeepCNet$(5,10)$, for a very long time, 1000 repetitions of the training set, gave a test error of 0.58\%. Using a GeForce GTX 680 graphics card, we can classify 3000 characters per second.


We tried increasing the number of hidden units. Training DeepCNet$(5,30)$ for 200 repetitions of the training set gave a test error of 0.46\%.

Dropout, in combination with increasing the number of hidden units and the training time generally improves ANN performance \cite{dropout}. DeepCNet$(5,60)$ has seven layers of matrix-multiplication. Dropping out 50\% of the input to the fourth layer and above  during training resulted in a test error of 0.31\%.

\section{The sparse signature grid representation}
The expression of the information contained in a path in the form of iterated integrals was pioneered by K.T. Chen \cite{Chen1958}. More recently, path signatures have been used to solve differential equations driven by rough paths \cite{LyonsQianRoughPathsBook,LyonsStFlour}.
The signature extracts enough information from the path to solve any linear differential equation and uniquely characterizes paths of finite length \cite{HamblyLyonsUniqueness}.

The signature has been used in sound compression \cite{soundCompressionRoughPaths}. A stereo audio recording can be seen as a highly oscillating path in $\RR^2$. Storing a truncated version of the path signature allows a version of the audio signal to be reconstructed.

Although computing the signature of a path is easy, the inverse problem is rather more difficult. The limiting factor in \cite{soundCompressionRoughPaths} was the lack of an efficient algorithm for reconstructing a path from its truncated signature when $m>2$. We side-step the inverse problem by learning to recognize the signatures directly.

\subsection{Computation of the path signature}
Let $[S,T]\subset \RR$ denote a time interval and let $V=\RR^d$ with $d=2$ denote the writing surface. Consider a pen stroke: a continuous function $X:[S,T]\to V$. For positive integers $k$ and intervals $[s,t]\subset[S,T]$, the $k$-th iterated integral of $X$ is the $d^k$-dimensional vector (i.e. a tensor in $V^{\otimes k}$) defined by
\[
X^k_{s,t}=\int_{s<u_1<\dots<u_k<t} 1\, dX_{u_1}\otimes\dots \otimes dX_{u_k}.
\]
By convention, the $k=0$ iterated integral is simply the number one. The $k=1$ iterated integral is the displacement of the path. The $k=2$ iterated integral is related to the curvature of the path.

As $k$ increases, it is a case of diminishing returns. The iterated integrals increase rapidly in dimension whilst carrying less information about the large scale shape of $X$. We therefore consider truncated signatures. The {\em signature}, truncated at level $m$, is the collection of the iterated integrals,
\[
S(X)_{s,t}=(1,X^1_{s,t},X^2_{s,t},\dots,X^m_{s,t}).
\]
With $d=2$, the dimension of this object is
\begin{align}\label{Mm}
M:=1+d+\dots+d^m=2^{m+1}-1.
\end{align}
Let $\Delta_{s,t}:=X_t-X_s$ denote the path displacement. Thinking of $\Delta_{s,t}$ as a row vector, the tensor product corresponds to the Kronecker matrix product ({\tt kron} in MATLAB).
When $X$ is a straight line, the signature can be calculated exactly:
\begin{align}\label{line-sig}
X^1_{s,t}=\Delta_{s,t},\q  X^2_{s,t}=\frac{\Delta_{s,t}\otimes\Delta_{s,t}}{2!},\q
X^3_{s,t}=\frac{\Delta_{s,t}\otimes\Delta_{s,t}\otimes\Delta_{s,t}}{3!},\q \dots
\end{align}
\begin{figure}
\begin{center}
\resizebox{14cm}{!}{\includegraphics{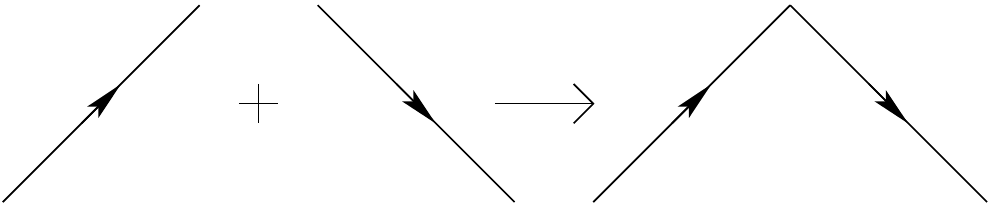}}
\resizebox{14cm}{!}{
$1\otimes(1,1)\otimes(\tfrac12,\tfrac12,\tfrac12,\tfrac12)$\q\q $1\otimes(1,-1)\otimes(\tfrac12,-\tfrac12,-\tfrac12,\tfrac12)$\qq\qq\q\q\ \
$1\otimes(2,0)\otimes(2,-1,1,0)$\qq\q\ \ }
\caption{Concatenating paths and the corresponding $m=2$ signatures.\label{sig-comp}}
\end{center}
\end{figure}
Joining together two paths to form a longer path results in a type of convolution of the signatures; see Figure~\ref{sig-comp}. With $s<t<u$, $S(X)_{s,u}$ is calculated from $S(X)_{s,t}$ and $S(X)_{t,u}$ using the formula
\begin{align}\label{sig-compose}
X^k_{s,u}=\sum_{i=0}^k X^i_{s,t}\otimes X^{k-i}_{t,u}, \q k=0,1,2,\dots,m.
\end{align}
Using \eqref{sig-compose} inductively we can calculate the signature for any piecewise linear  path.

\subsection{Representing pen strokes}
Each character collected by an electronic stylus is represented by a sequence of pen strokes. Each stroke is represented by a sequence of points that we will treat as a piecewise linear path.

Recall that $n\approx N/3$ denotes the scale at which characters will be drawn. We use another parameter $\delta=n/5$ to describe very approximately the scale of path curvature.

Here is an algorithm that takes a character and uses the first $m$ iterated integrals to construct an $N\times N\times M$ CNN-input-layer array \eqref{Mm}.
\begin{dashlist}
\item Initialize an array of size $N\times N\times M$ to all zeros. Think of this as an $N\times N$ array of vectors in $\RR^M$; the first two dimensions correspond to the writing surface, and the third corresponds to the elements of the signature.
\item Rescale the character to fit in an $n\times n$ box placed in the center of the $N\times N$ input layer. Let $X_i:[0,\mathrm{length}(i)]\to[0,N]^2$ denote the $i$-th character stroke, parameterized to have unit speed.

\item Let $X_i(t)$ denote a point moving along the $i$-th stroke. Mapping $X_i(t)$ into the $N\times N$ grid, store $S(X_i)_{t-\delta,t+\delta}$ in the appropriate column of the array.
\end{dashlist}
Note that the first element of the signature corresponds to the zero-th iterated integral which is always a one. Thus if we look at the first $N\times N \times 1$ slice of our 3D array we see a 2D bitmap picture of the character. If $m\ge1$, the next two layers contain the first iterated integrals: they encode the direction the pen was moving. If $m\ge2$, the next four layers encode the second iterated integrals, and so on.

\section{Results}
\subsection{10, 26 and 183 character classes}
We will first look at three relatively small datasets \cite{UCIrep} to study the effect of increasing the signature truncation level $m$.
\begin{dashlist}
\item The Pendigits dataset contains handwritten digits 0-9. It contain about 750 training characters for each of the ten classes.
\item The UJIpenchars database includes the 26 lowercase letters. The training set contains 80 characters for each of the 26 classes.
\item The Online Handwritten Assamese Characters Dataset contains 45 samples of 183 Indo-Aryan characters. We used the first $k$ handwriting samples as the training set, and the remaining $45-k$ samples for a test set ($k=15$ or 36).
\end{dashlist}
To make the comparisons interesting, we deliberately restrict the DeepCNet $k$ and $l$ parameters. The justification for this is that increasing $k$ and $l$ is computationally expensive. In contrast, increasing $m$ only increases the cost of evaluating the first layer of the CNN; in general for sparse CNNs the first layer represents only a small fraction of the total cost of evaluating the network. Thus increasing $m$ is cheap, and we will see that it tends to improve test performance.

We tried smaller and larger networks, and using the training set with and without increasing its size by affine transformations (a random mix of scalings, rotations and translations). The table shows the test set error rates.

\hspace{-1cm}\begin{tabular}{|cccc|cccc|}
\hline
Dataset & DeepCNet & $n$ & Transforms & $m=0$ & $m=1$ & $m=2$ & $m=3$ \\
\hline
Pendigits & (3,10) & 10 & $\times$ & 3.37\% & 1.60\% & 1.32\% & 1.09\%\\
Pendigits & (5,50) & 32 & $\checkmark$ & 0.94\% & 0.43\% & 0.40\% & 0.40\%\\
UJI lowercase & (4,10) & 20 & $\times$ & 18.3\% & 16.3\% & 15.3\% & 13.4\%\\
UJI lowercase & (5,50) & 32 & $\checkmark$ & 7.4\% & 6.6\% & 6.9\% & 6.9\%\\
Assamese-15 & (5,20) & 32 & $\times$ & 48.9\% & 40.3\% & 39.8\% & 34.8\%\\
Assamese-15 & (5,50) & 32 & $\checkmark$ & 12.5\% & 11.9\% & 11.0\% & 11.0\%\\

Assamese-36 & (6,50) & 64 & $\checkmark$ & 2.2\% & 2.3\% & 1.6\% & 2.3\%\\
\hline
\end{tabular}

\noindent The results show that the first and second, and even the third iterated integrals carry information that CNNs can use to improve generalization from the training to the test set.

\subsection{CASIA}
The CASIA-OLHWDB1.1\cite{CASIA} database contains samples from 300 writers, allocated into training and test sets. It contains samples of the 3755 GBK level-1 Chinese characters.

A test error of 5.61\% is achieved using a deep CNN applied to pictures drawn from the pen data \cite{multicolumndeep}. Their program took advantage of a couple of features that are often used in the context of CNNs. Rather than simply drawing a binary bitmap of the character, they convolved the images with a Gaussian blur. They also used elastic distortions.

We trained a DeepCNet$(6,100)$ with $m=2$ and $n=60$.
We went through the training data 80 times. For the first 40 repetitions, we randomized the placement of the training characters is a small neighborhood of the center of the input layer. This gave a test error of 4.44\%. We then applied affine transformations to the training characters for another 40 repetitions, giving a test error of 4.01\%.

Modifying the above network by adding dropout---with dropout per level of $0,0,0,0.1,0.2,0.3,0.4,0.5$---resulted in a test error to 3.58\%.

\section{Conclusion}
We have studied two methods for improving the performance of CNNs for online handwriting character recognition, enhancing the pictures with signature information and using sparsity to increase the depth of the networks we can train.
They work well together on a variety of alphabets.

This work raises a number of questions:
\begin{dashlist}
\item Besides the path signature, there are many other ways of describing the shape of a path. You could calculate 8-direction histograms \cite{bb104561} to give a sparse $N\times N\times 8$ input layer. Or you could use an unsupervised learning algorithm to characterize path segments. What is the best way of describing paths for CNNs?

\item Can our approach be applied to natural images using curves extracted from the image by some deterministic curve tracing algorithm?

\item Our CNN is sparse with respect to the two spatial dimensions, but not in terms of the third feature-set dimension. Predictive Sparse Decomposition \cite{
conf/nips/RanzatoPCL06} results in sparse feature vectors. Can doubly-sparse CNNs be built to recognize images more efficiently?

\end{dashlist}
\section{Acknowledgement}
Many thanks to Fei Yin, Qiu-Feng Wang and Cheng-Lin Liu for their hard work organizing the ICDAR2013 competition.


\end{document}